\documentclass[10pt]{cai}

\graphicspath{{figures/}}

\usepackage[utf8]{inputenc} 
\usepackage[T1]{fontenc}    
\usepackage{hyperref}       
\usepackage{url}            
\usepackage{booktabs}       
\usepackage{amsfonts}       
\usepackage{nicefrac}       
\usepackage{microtype}      
\usepackage{xcolor}         

\usepackage{makecell}
\usepackage{mathtools,xparse}
\usepackage{amsmath}
\usepackage{amsthm}
\usepackage{graphicx}
\usepackage{xcolor}
\usepackage{algorithm}
\usepackage{algorithmicx}
\usepackage{tikz,pgfplots,pgf}
\usetikzlibrary{matrix,shapes,arrows,positioning}
\usepackage{subcaption}
\usepackage{soul}

\usepackage{definitions}

\usepackage{listings}
\usepackage[noend]{algpseudocode}

\usepackage{multirow}

\pgfplotsset{compat=1.17}

\begin{document}

\volumeheader{36}{0}
\begin{center}

  \title{PAC-Bayesian Learning of Aggregated Binary Activated Neural Networks with Probabilities over Representations}
  \maketitle

  \thispagestyle{empty}

  \begin{tabular}{cc}
    Louis Fortier-Dubois\upstairs{*}, Benjamin Leblanc, Gaël Letarte, François Laviolette, Pascal Germain
\\[0.25ex]
   {\small Département d'informatique et de génie logiciel, Université Laval} \\
  \end{tabular}

  \emails{
    \upstairs{*}louis.fortier-dubois.1@ulaval.ca
    }
  \vspace*{0.2in}
\end{center}

\begin{abstract}
Considering a probability distribution over parameters is known as an efficient strategy to learn a neural network with non-differentiable activation functions. 
We study the expectation of a probabilistic neural network as a predictor by itself, focusing on the aggregation of binary activated neural networks with normal distributions over real-valued weights. 
Our work leverages a recent analysis derived from the PAC-Bayesian framework that derives tight generalization bounds and learning procedures for the expected output value of such an aggregation, which is given by an analytical expression. While the combinatorial nature of the latter has been circumvented by approximations in previous works, we show that the exact computation remains tractable for deep but narrow neural networks, thanks to a dynamic programming approach. 
This leads us to a peculiar bound minimization learning algorithm for binary activated neural networks, where the forward pass propagates probabilities over representations instead of activation values.
A stochastic counterpart that scales to wide architectures is proposed.\\
\end{abstract}

\copyrightnotice

\section{Introduction}
\label{section:introduction}

The computation graphs of deep neural networks (\emph{a.k.a.} architectures) are challenging to analyze, due to their multiple composition of non-linearities \citep{caruana2014}, overparametrization \citep{Zhang2017} and highly non-convex learning objective \citep{dauphin2014,choromanska2015}. Studying simpler models seems a sensible strategy to gain insight on neural network behaviour and state performance guarantees \mbox{\citep[\eg][]{arora2019,belkin2020}}. Our work starts from one possible simplification, obtained by considering the binary activation function, meaning that each neuron outputs only one bit of information instead of the many bits needed to represent a real number. 
The use of neural networks involving binary weights, with or without binary activation \citep{courbariaux2015binaryconnect,hubara2016,soudry2014expectation}, has been suggested for reducing their resource consumption, and these may be especially useful in view of using a pre-trained network for forward propagation on embedded systems, but this is not our primary objective. 
We aim to foster an atypical vision of neural networks, where binary activated networks with real-valued parameters are viewed as elementary pieces of an ensemble, that we study as a whole.

Our work is motivated by the analysis of \citet{letarte2019dichotomize,biggs2021}, rooted in the PAC-Bayes theory~\citep{mcallester-99}: they propose a learning objective for aggregation of binary activated networks derived from a high-confidence generalization bound, and they empirically show that minimizing this objective provides a predictor with both tight and theoretically sound guarantees. They also derived an analytical expression for the expected output value of binary activated networks sampled from Gaussian distributions. Being differentiable, this expression enables gradient descent optimization, and is to be added to methods to train networks with non-differentiable activation functions~\mbox{\citep{Williams1992,bengio13,hubara2017quantized}}. 
However, in order to preserve valid PAC-Bayes guarantees and be able to train large neural networks, Letarte et al.~\cite{letarte2019dichotomize} rely on an approximation: neurons on the same layer all treat their inputs as if they were independent draws of the same probability distribution, when in fact all the inputs must correspond to the same draw.\footnote{This recalls the \emph{mean-field approximation} performed by \citet{soudry2014expectation} for learning binary activated networks in a Bayesian setting.} Not only does this make the algorithm output values that are slightly straying from the true aggregation expectation, it also fatally increases the PAC-Bayes bound for deeper architectures. 
Lately, \citet{biggs2021} revisited the PAC-Bayes aggregation of neural networks, notably providing lower-variance approximation schemes; doing so, they exhibited that the \emph{forward propagation} function of Letarte et al. can be rewritten in order to remove the above-mentioned independence assumption. 
Starting from this result, we conduct the major part of our study stepping away from all approximations.

Hence, our first contribution highlights that the exact expectation of an aggregation of binary activated networks is computable in time exponential in the width of the network, and linear in its depth. The originality of our proposed learning algorithm relies on computing probabilities of occurrences of the hidden layer representations. Doing so, not only our algorithm allows us to obtain non-vacuous PAC-Bayes bounds for very deep binary activated networks, but reveals itself as an interesting prediction mechanism. Noteworthy, we show that once the parameters are optimized, the prediction on a new example is achievable with a time complexity that remains \textit{constant} relatively to the network depth. 
Our second contribution consists in the analysis of a compact scheme of our resulting predictors, having a constant computing time regarding the network depth, and the dichotomy between the first layer of the network versus the following layers.
For both these contributions, we present a stochastic version with sub-exponential time complexity regarding the network width.

\section{Background and notation}
\label{section:background}

\tikzset{%
	every neuron/.style={
		circle,
		draw,
		minimum size=0.2cm
	},
	neuron missing/.style={
		draw=none, 
		scale=2,
		text height=0.3cm,
		execute at begin node=\color{black}$\vdots$
	}
}

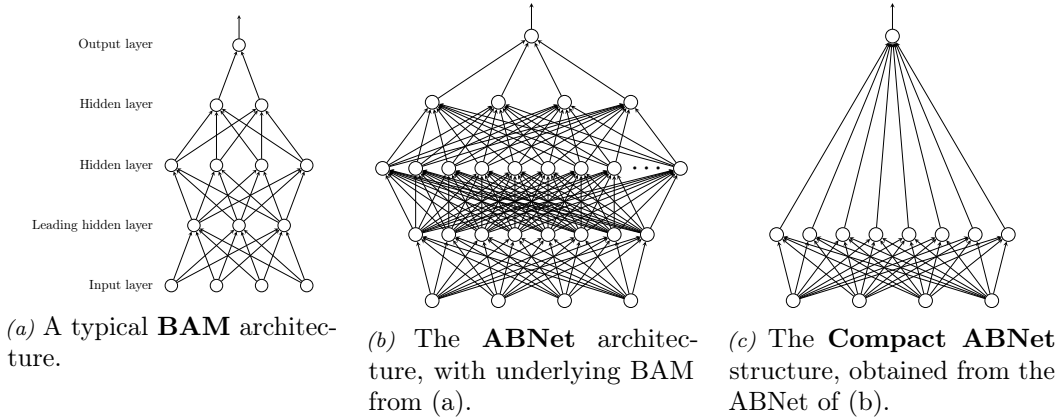
\begin{figure*}[t]
    \centering
     \begin{subfigure}{0.31\textwidth}
     \centering
        \rotatebox{90}{\scalebox{.5}{\tikzset{%
	every neuron/.style={
		circle,
		draw,
		minimum size=0.2cm
	},
	neuron missing/.style={
		draw=none, 
		scale=2,
		text height=0.3cm,
		execute at begin node=\color{black}$\vdots$
	}
}

\begin{tikzpicture}[x=0.8cm, y=1.2cm, >=stealth]
    	
    	
    	\foreach \m [count=\y] in {1,2,3,4}
    	\node [every neuron] (data-\m) at (0,2.5-\y) {};
    	
    	\foreach \m [count=\y] in {1,2,3}
    	\node [every neuron] (input-\m) at (2,2-\y) {};
    	
    	\foreach \m [count=\y] in {1,2,3,4}
    	\node [every neuron] (hidden1-\m) at (4,2.5-\y) {};
    	
    	\foreach \m [count=\y] in {1,2}
    	\node [every neuron] (hidden2-\m) at (6,1.5-\y) {};	
    	
    	\foreach \m [count=\y] in {1}
    	\node [every neuron] (output-\m) at (8,0) {};
    	
    	
    	\foreach \i in {1,...,4}
    	\foreach \j in {1,...,3}
    	\draw [->] (data-\i) -- (input-\j);
    	
    	\foreach \i in {1,...,3}
    	\foreach \j in {1,...,4}
    	\draw [->] (input-\i) -- (hidden1-\j);
    	
    	\foreach \i in {1,...,4}
    	\foreach \j in {1,...,2}
    	\draw [->] (hidden1-\i) -- (hidden2-\j);
    	
    	\foreach \i in {1,...,2}
    	\foreach \j in {1,...,1}
    	\draw [->] (hidden2-\i) -- (output-\j);
    	
    	\foreach \l [count=\i] in {1}
    	\draw [->] (output-\i) -- ++(1,0)
    	node [above, midway] {};
    	
    \foreach \l [count=\x from 0] in {\rotatebox{270}{Input layer}, \rotatebox{270}{Leading hidden layer}, \rotatebox{270}{Hidden layer}, \rotatebox{270}{Hidden layer}, \rotatebox{270}{Output layer}}
\node [align=center, above] at (\x*2,1.8) {\l};
    	
    \end{tikzpicture}}}
        \caption{\normalsize A typical \textbf{BAM} architecture.\\}
        \label{fig:base_nn_}
    \end{subfigure}
    \quad
    \begin{subfigure}{0.31\textwidth}
    \centering
        \rotatebox{90}{\scalebox{.55}{\tikzset{%
	every neuron/.style={
		circle,
		draw,
		minimum size=0.2cm
	},
	neuron missing/.style={
		draw=none, 
		scale=2,
		text height=0.3cm,
		execute at begin node=\color{black}$\vdots$
	}
}

\begin{tikzpicture}[x=0.8cm, y=0.8cm, >=stealth]
	
	
	\foreach \m [count=\y] in {1,2,3,4}
	\node [every neuron] (data-\m) at (0,9.5-2*\y) {};
	
	\foreach \m [count=\y] in {1,2,3,4,5,6,7,8}
	\node [every neuron] (input-\m) at (2,9-\y) {};
	
	\foreach \m [count=\y] in {1,2,3,4,5,6,7,8,missing,9}
	\node [every neuron/.try, neuron \m/.try] (hidden1-\m) at (4,10-\y) {};
	
	\foreach \m [count=\y] in {1,2,3,4}
	\node [every neuron] (hidden2-\m) at (6,9.5-2*\y) {};	
	
	\foreach \m [count=\y] in {1}
	\node [every neuron] (output-\m) at (8,4.5) {};
	
	
	\foreach \i in {1,...,4}
	\foreach \j in {1,...,8}
	\draw [->] (data-\i) -- (input-\j);
	
	\foreach \i in {1,...,8}
	\foreach \j in {1,...,9}
	\draw [->] (input-\i) -- (hidden1-\j);
	
	\foreach \i in {1,...,9}
	\foreach \j in {1,...,4}
	\draw [->] (hidden1-\i) -- (hidden2-\j);
	
	\foreach \i in {1,...,4}
	\foreach \j in {1,...,1}
	\draw [->] (hidden2-\i) -- (output-\j);
	
	\foreach \l [count=\i] in {1}
	\draw [->] (output-\i) -- ++(1,0)
	node [above, midway] {};
	
\end{tikzpicture}}}
        \caption{\normalsize  The \textbf{ABNet} architecture, with underlying BAM from~(a).}
        \label{fig:abnet_}
    \end{subfigure}
    \quad
    \begin{subfigure}{0.31\textwidth}
    \centering
       \rotatebox{90}{\scalebox{.55}{\tikzset{%
	every neuron/.style={
		circle,
		draw,
		minimum size=0.2cm
	},
	neuron missing/.style={
		draw=none, 
		scale=2,
		text height=0.3cm
	}
}

\begin{tikzpicture}[x=0.8cm, y=0.8cm, >=stealth]
	
	
	\foreach \m [count=\y] in {1,2,3,4}
	\node [every neuron] (data-\m) at (0,9.5-2*\y) {};
	
	\foreach \m [count=\y] in {1,2,3,4,5,6,7,8}
	\node [every neuron] (input-\m) at (2,9-\y) {};

	\foreach \m [count=\y] in {1}
	\node [every neuron] (output-\m) at (8,4.5) {};
	
	
	\foreach \i in {1,...,4}
	\foreach \j in {1,...,8}
	\draw [->] (data-\i) -- (input-\j);
	
	\foreach \i in {1,...,8}
	\foreach \j in {1,...,1}
	\draw [->] (input-\i) -- (output-\j);
	
	\foreach \l [count=\i] in {1}
	\draw [->] (output-\i) -- ++(1,0)
	node [above, midway] {};
	
\end{tikzpicture}}}
       \caption{\normalsize  The \textbf{Compact ABNet} structure, obtained from the ABNet of (b).}
       \label{fig:compact_abnet_}
    \end{subfigure}
    \newline
    \caption{The BAM shown in (\subref{fig:base_nn_}) has depth $L=4$, and widths 
    $\langle d_k \rangle_{k=0}^4 = \langle 4,3,4,2,1 \rangle$.
    In (\subref{fig:abnet_}), hidden layers have a width of $2^{d_k}$ (\ie, one neuron per binary representation obtainable from a layer width of $d_k$ in the underlying BAM).
    In (\subref{fig:compact_abnet_}), all hidden layers are merged, leaving a depth of $2$ (as explained in Section~\ref{section:compact}).}
\end{figure*}

We focus our study on the task of binary classification, using \textit{binary activated multilayer} (BAM) neural networks, \ie, networks where each neuron either outputs $-1$ or $+1$, using the sign function: $\sgn (x) = -1$ if $x < 0$, $+1$ otherwise.
We consider fully connected BAMs of $L \in \mathbb{N}^+$ layers of size $d_k \in \mathbb{N}^+, \; \forall k \in \{1, 2, \dots, L\}$, and inputs of size $d_0 \in \mathbb{N}^+$. We fix $d_L = 1$, whose output is the classification output of the whole network. We call $L$ the \textit{depth} of the network, $d_k$ the \textit{width} of the $k$\ts{th} layer, and 
$\max_{k \in \{1, 2, \dots, L\}}{d_k}$
the \textit{width} of the network. The sequence $\langle d_k \rangle_{k=0}^L$ constitutes a (fully connected) \textit{architecture}. Unlike in \textit{binary} neural networks \cite[\eg][]{courbariaux2015binaryconnect}, the parameters of a BAM are not constrained to be binary, but only activations are binary-valued. We thus have weights $\mathbf{W}_k \in \mathbb{R}^{d_k \times d_{k-1}}$ and biases $\mathbf{b}_k \in \mathbb{R}^{d_k}$, for $k \in \{1, 2, \dots, L\}$. That being said, in the remaining, the equations will be stated without loss of generality in terms of the weights $\mathbf{W}_k$ only. Therefore, a BAM $\mathcal{B}$ is totally defined by the tuple $\Bcal \eqdots \langle \Wbf_k \rangle_{k=1}^{L}$. See Fig.~\ref{fig:base_nn_} for an example of architecture. The $0$\ts{th} layer is the input layer, the $1$\ts{st} one is the \textit{leading} hidden layer, any $k$\ts{th} layer, with $1 < k < L$, is simply called a hidden layer and the $L$\ts{th} one is the output layer. Following the premises of Letarte et al. \cite{letarte2019dichotomize}, we consider a distribution over BAMs, which we call an aggregation of BAMs. 

\begin{definition}
An \emph{aggregation of BAMs} with mean parameters $\mathcal{B}_M$, denoted $\mathcal{A}(\mathcal{B}_M)$, 
is given by an isotropic Gaussian probability distribution over all parameters, centered in $\mathcal{B}_M$. 
\end{definition}

The BAM forward propagation process consists of computing the following function for an input $\mathbf{x} {\,\in\,} \mathbb{R}^{d_0}$, where $\sgn$ is defined element-wise:
$$F_{\mathcal{B}}(\mathbf{x}) = \sgn (\mathbf{W}_L \sgn (\mathbf{W}_{L-1} \sgn (\dots \sgn(\mathbf{W}_1 \mathbf{x}) \dots)) )\,.$$
In other words, the output of a BAM $\Bcal$ is $F_{\mathcal{B}}(\mathbf{x}) \eqdots F_{\mathcal{B}}^L(\mathbf{x})$, given the recursive equations
\begin{align*}
    F_{\mathcal{B}}^k(\mathbf{x})
    =
    \begin{cases}
        \sgn(\mathbf{W}_1 \mathbf{x}) & \mbox{if $k=1$},\\[1mm]
        \sgn(\mathbf{W}_k F_{\mathcal{B}}^{k-1}(\mathbf{x})) & \mbox{otherwise}.
    \end{cases}
\end{align*}
Beside, the output of an \emph{aggregation of BAMs} is the expected output of a BAM drawn from the (continuous) parameters distribution:
\begin{align} \label{eq:FABM}
F_{\Acal(\Bcal_M)}(\xbf) = \textstyle\Esp_{\Bcal \sim \Acal(\Bcal_M)} F_{\Bcal}(\xbf).
\end{align}
Such aggregation of BAMs is our main object of study. Note that the forthcoming Section~\ref{section:pacbayes} is dedicated to a PAC-Bayesian treatment where $\Acal(\Bcal_M)$ is considered as a \emph{posterior distribution}. In line with the \mbox{(PAC-)}Bayesian literature, we call the single BAM network $\mathcal{B}_M$ the \textit{Maximum-A-Posteriori} (MAP) predictor.



\section{The aggregation output}
\label{section:dynamic_programming}

We present a recursive formulation to compute the exact aggregation output, denoted $F_{\Acal(\Bcal_M)}(\xbf)$, which is differentiable end-to-end. 
The formulation is equivalent to the one of \citet[Eq. (9)]{biggs2021}, but is expressed in order to highlight the probabilities of representations for each layer, which is the cornerstone of our analysis.


Given a \emph{fixed} representation (an input vector $\abf \in \Rbb^d$), the expected output of a single neuron with sign activation over an isotropic Gaussian distribution centered on $\wbf\in\Rbb^d$ is
\begin{align*}
   \Esp_{\mathbf{v} \sim \mathcal{N}(\mathbf{w}, \mathbf{I})} \sgn(\mathbf{v} \cdot \mathbf{a}) = \Erf\bigg(\frac{\mathbf{w} \cdot \mathbf{a}}{\sqrt{2} \lVert \mathbf{a} \rVert}\bigg)\,,
\end{align*}
where $\Erf(x) = \frac{2}{\sqrt{\pi}}\int_0^x e^{-t^2} dt$ is the Gauss error function. However, as the input vector of a neuron relies on the distribution of weights on the previous layers, the representation itself is a random vector.
A neuron outputs a value $s\in \{-1,1\}$ with probability

\begin{align} \label{eq:prob_1neuron}
    \Pr(f_\wbf(\abf) = s) =  \frac{1}{2} + \frac{s}{2} \Erf\!\left(\frac{\wbf \cdot \abf}{\sqrt{2} \lVert \abf \rVert}\right).
\end{align}
Thus, the probability of observing a specific representation $\sbf = (\sbf^1, \ldots\,, \sbf^{d_k}) \in R_k$ (with $R_k \eqdots \{-1,1\}^{d_k}$) at layer $k$ can be expressed in a recursive manner, having the events $a_{k}^{\sbf} \eqdots F_{\mathcal{B}}^k(\mathbf{x}) = \sbf$ and $a_{k-1}^{\bar\sbf} \eqdots F_{\mathcal{B}}^{k-1}(\mathbf{x}) = \bar\sbf$~:
\begin{align*}
\Pr(a_{k}^{\sbf}) 
=
\begin{cases}
\dprod_{i=1}^{d_1} \bigg(\frac{1}{2} + \frac{\sbf^i}{2} \Erf\!\left(\frac{\mathbf{W}_{1}^i \cdot \mathbf{x}}{\sqrt{2} \lVert \mathbf{x} \rVert}\right)\bigg) & \mbox{if $k=1$},\\[4mm]
\dsum_{\bar\sbf \in R_{k-1}} 
\Pr(a_{k}^{\sbf} \given a_{k-1}^{\bar\sbf}) \Pr(a_{k-1}^{\bar\sbf}) & \mbox{otherwise}.
\end{cases}
\end{align*}
The base case of the above recursion refers to the leading hidden layer ($F_{\mathcal{B}}^1$). The probability of observing a representation $\sbf \in R_1$ on this first hidden layer given an input~$\xbf$ amounts to be the product of the probability associated with the $d_1$ individual neuron values (Eq.~\ref{eq:prob_1neuron}).
The general case refers to the subsequent hidden layers ($F_{\mathcal{B}}^2, \ldots, F_{\mathcal{B}}^L$). The probability of observing a representation $\sbf \in R_k$ is decomposed into the sum of the $2^{d_{k-1}}$ probabilities of observing the representation $\bar\sbf \in R_{k-1}$ on the previous hidden layer (obtained recursively) and the conditional probability
(obtained from Eq~\eqref{eq:prob_1neuron}, using the fact that $\|\bar\sbf\|^2 = d_{k-1}$) :
\begin{align*}
 \Pr(a_{k}^{\sbf} \given a_{k-1}^{\bar\sbf}) = \prod_{i=1}^{d_k} \left[ \frac{1}{2} + \frac{\sbf^i}{2} \Erf\!\left(\frac{\mathbf{W}_{k}^i \cdot \bar\sbf}{\sqrt{2 d_{k-1}}}\right)  \right]\,.
\end{align*} 
Finally, assuming the output layer is only one neuron wide, the exact output of the aggregation can be computed with:
$F_{\Acal(\Bcal_M)}(\xbf) = \Esp_{\Bcal \sim \Acal(\Bcal_M)}  F_{\Bcal}(\xbf) = \Pr( F_{\Bcal}^{L}(\xbf) {=} 1) - \Pr( F_{\Bcal}^{L}(\xbf) {=} -1).$

By contrast, the proposed PBGNet algorithm of Letarte et al.~\cite[Eq.~(16)]{letarte2019dichotomize} computes $F_{PBG}^L(\xbf)$ using the following equations (equivalent to the ones above only when $L\leq2$):
\begin{align*}
F_{PBG}^k(\xbf) = 
\begin{cases}
\Erf\!\left(\frac{\mathbf{W}_{1} \cdot \xbf}{\sqrt{2} \lVert \xbf \rVert}\right) & \mbox{if $k=1$},\\
\dsum_{\bar\sbf \in R_{k-1}} 
\Erf\!\left(\tfrac{\mathbf{W}_{k} \cdot \bar\sbf}{\sqrt{2 d_{k-1}}}\right) \dprod_{i=1}^{d_k}
\left[\frac{1}{2} + \frac{\bar\sbf^i}{2} \left(F_{PBG}^{k-1}(\xbf)\right)^i\right]
& \mbox{otherwise}.
\end{cases}
\end{align*}
This comes down to outputting at each layer only the expectation of the BAM representation given the expectation of the previous layer, instead of our complete probability distribution. Their method therefore deletes information at each layer since the expectation does not carry the correlation between each individual neuron output. This approximation recalls the mean-field one, on which the Bayesian analysis of binary activated networks of \citet{soudry2014expectation} relies. We avoid such approximations, and we compute $F_{\Acal(\Bcal_M)}(\xbf)$ by a dynamic programming approach, described next.

\textbf{Dynamic program.} \ 
Abusing notation a little, when writing $\big[ g(\sbf) \big]_{\sbf \in R}$ we assume all $\sbf$ are taken in lexicographical order.
Hence, posing
\begin{align} \label{eq:psi_k}
    \mathbf{\Psi}_k = \bigg[\prod_{i=1}^{d_k} \bigg(\frac{1}{2} + \frac{\sbf^i}{2} \Erf\big(\frac{\mathbf{W}_{k}^i \cdot \bar\sbf}{\sqrt{2 d_{k-1}}}\big)\bigg)\bigg]_{\sbf \in R_{k},\bar\sbf\in R_{k-1}},
\end{align}
one can obtain straightforwardly the probability vector
$\mathbf{P}_{k} = \big[\Pr(F_\mathcal{B}^{k}(\mathbf{x}) \sbf)\big]_{\sbf \in R_{k}}$  by computing $\mathbf{\Psi}_k \cdot \mathbf{P}_{k-1}\,$. Starting with 
\begin{align} \label{eq:p1_x}
    \mathbf{P}_1(\xbf) = \bigg[\prod_{i=1}^{d_1}  \bigg(\frac{1}{2} + \frac{\sbf^i}{2} \Erf\big(\frac{\mathbf{W}_{1}^i \cdot \mathbf{x}}{\sqrt{2} \lVert \mathbf{x} \rVert}\big)\bigg)\bigg]_{\sbf \in R_1}\,,
\end{align} and computing $\mathbf{P}_k$ for $k \in \{2, 3, \dots, L\}$ in ascending order, we can therefore compute the exact expectation of a BAM $\mathcal{B} \sim \mathcal{A}$ in time exponential in $\mathcal{B}$'s width, yet linear in its depth. 



The previous formulas lead to what stands as the forward propagation process of our new neural network, which we name \textit{ABNet} for \textit{\textbf{A}ggregation of \textbf{B}inary activated \textbf{Net}works}. From parameters $\mathcal{B}_M$, it computes $F_{\Acal(\Bcal_M)}(\xbf)$. The computation graph of ABNet is illustrated by Fig.~\ref{fig:abnet_}.  The width of hidden layers $k < L$ are of exponential size $2^{d_k}$ relatively to the width $d_k$ of the BAM networks it aggregates. Each layer of ABNet outputs a probability distribution over all possible configurations of the underlying BAM. The next layer then multiplies those probabilities by the conditional probabilities $\mathbf{\Psi}$, which is just a reorganization of the weights and is totally independent of the input $\mathbf{x}$. As a result, ABNet applies only linear functions on hidden and output layers; an observation we discuss further in Section \ref{section:compact}. 

\textbf{Stochastic version.} \ 
ABNet has many interesting theoretical properties, but the necessity of computing the probability of every combination of neuron outputs at a given layer makes it too cumbersome for practical applications. We propose a stochastic version of ABNet, which keeps its property of avoiding the mean field approximation while limiting the computation complexity with regard to the width to a quadratic one. Note that the stochastic versions of ABNet and PBGNet are truly different: while the former propagates probabilities over representations, the latter relies on forward and backward passes alike standard neural networks. This is achievable by picking a constant number $n$ of representations $R'_k$ uniformly from $R_k$ at layer $k$, computing only the occurring probability of those $n$ representations (replacing representation sets $R_k$ by their uniformly drawn counterparts $R'_k$ in Equations \ref{eq:psi_k} and \ref{eq:p1_x}), and normalizing at each layer by dividing $\Pbf_k$ by $\sum_{\sbf_k \in R'_k} \Pbf_k[\sbf_k]$. 

\begin{figure*}[t]
     \begin{subfigure}[t]{0.48\textwidth}
        \centering
        \includegraphics[scale=0.48]{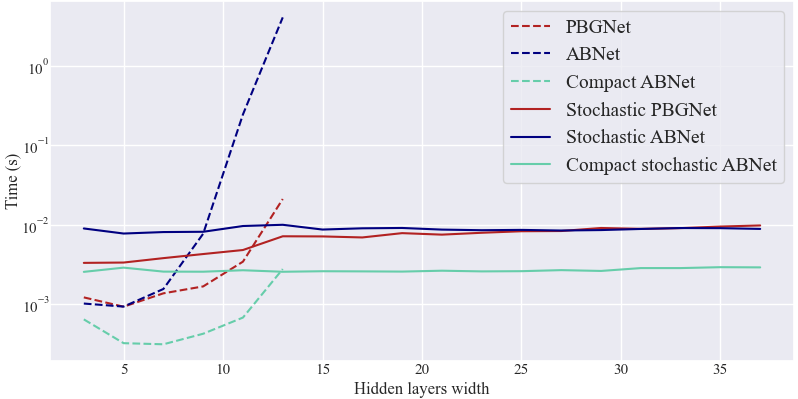}
        \caption{
        Computation time according to the network width, with depth $L=6$. }
        \label{fig:varying_width}
    \end{subfigure}
    \quad
     \begin{subfigure}[t]{0.48\textwidth}
        \centering
        \includegraphics[scale=0.48]{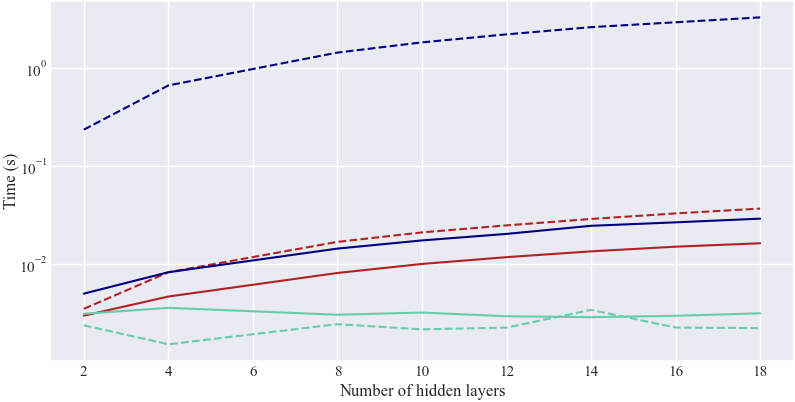}
        \caption{
        Computation time according to the network depth, with width $d_k=12$ for $1 \leq k < L$. }
        \label{fig:varying_depth}
    \end{subfigure}
    \quad
    \caption{
    Empirical study of the time needed for the forward propagation of our four ABNet versions and the benchmark PBGNet, on a batch of 32 examples of the Ads dataset \citep{Dua:2019}, with 100 samples for stochastic versions, averaged on 100 repetitions. As can be seen in Fig.~\ref{fig:varying_width}, the memory requirements of our (non-stochastic) PBGNet and ABNet implementations exceed the available resources for layer widths greater than 13. 
    \label{fig:varying}
}
\end{figure*}

\textbf{Complexity.} \ Assuming every layer has the same width $d$ at each layer, the complexity of PBGNet is $\Ocal (L2^d d^2)$ (or $\Ocal (Lnd^2)$ for its stochastic counterpart, with $n$ samples), while the complexity of ABNet is $\Ocal (L2^{2d} d^2)$ (or $\Ocal (Ln^2 d^2)$ for the stochastic version). See Fig.~\ref{fig:varying} for an empirical study of computing times. A salient fact is that stochastic versions scale much better on wide architectures (Fig.~\ref{fig:varying_width}). 

\section{Bounding and optimizing the generalization loss}
\label{section:pacbayes}

Initiated by \citet{mcallester-99}, the PAC-Bayes theory allows one to bound the generalization error of a learned predictor without requiring a validation set, under the sole assumption that data is sampled in an $\iid$ way from the unknown distribution $\mathcal{D}$. 

To be eligible for a PAC-Bayesian treatment, predictors must be expressed through a \emph{posterior} probability distribution over a predefined class of \emph{hypotheses}. Even though neural networks are not naturally defined as such, many valuable analyses have been proposed by applying a PAC-Bayesian theory to stochastic variants of deterministic neural networks \citep{LangfordCaruana2001,DziugaiteRoy2018,zhou2018nonvacuous,perez-ortiz2020,pitas2020} by considering perturbations (typically Gaussian distributed noise) on the weights. This strategy can be applied to any neural network topology
and activation functions, but the generalization bounds do not apply to the underlying deterministic (non-perturbed) network. Other theoretical frameworks than PAC-Bayes also  leverage probabilistic views of neural networks, typically studying convergence of infinitely wide networks \cite{DBLP:journals/corr/abs-1711-00165,DBLP:conf/nips/JacotHG18}.

By adopting the construction of Letarte et al. \cite{letarte2019dichotomize} and designing our predictor \emph{natively} as a distribution over (finite) BAM networks (Eq~\ref{eq:FABM}), the PAC-Bayesian bound applies to the output of ABNet.
The forthcoming Theorem~\ref{thm:pacbayes} provides high confidence upper bound for the \emph{generalization loss} of a learned ABNet, defined as $
  \mathcal{L}_{\mathcal{D}}(F_{\mathcal{A}}) = \Esp_{(\mathbf{x},y) \sim \mathcal{D}} \; \ell(F_{\mathcal{A}}(\mathbf{x}), y),$
where $\ell(y', y) = \frac{1}{2}(1-yy') \in [0,1]$ is the linear loss for the binary classification problem. The two main quantities involved in the computation of the bound are the \emph{empirical} loss on the learning sample $S = \{(\mathbf{x}_i, y_i)\}_{i=1}^n \sim \mathcal{D}^n$,
\begin{equation}\label{eq:emploss}
    \hat{\mathcal{L}}_{S}(F_{\mathcal{A}}) = \frac{1}{n} \sum_{i=1}^n \ell(F_{\mathcal{A}}(\mathbf{x}_i), y_i)\,,
\end{equation}
 and the \emph{Kullback-Leibler (KL) divergence} between the learned parameters (posterior distribution) $\Bcal = \langle \Wbf_k \rangle_{k=1}^{L}$ and a reference (prior distribution) $\mathcal{B}^p = \langle \Wbf_k^p \rangle_{k=1}^{L}$ which is independent of the training data.\footnote{Following the common practice \cite{DziugaiteRoy2018,letarte2019dichotomize,biggs2021,pitas2020}, we chose an SGD random initialization as $\mathcal{B}^p$.} By using isotropic Gaussians for both the prior and the posterior, the KL divergence is easily obtained with
 \begin{equation}\label{eq:KL}
   \KL(\mathcal{B} \| \mathcal{B}^p) = \frac{1}{2} \sum_{k=1}^L \lVert \mathbf{W}_k - \mathbf{W}_k^p \rVert ^2 .  
 \end{equation}
The PAC-Bayes theorem below is borrowed from Letarte et al. \cite{letarte2019dichotomize}, which itself is a variation from a seminal result from \citet{catoni-07}, but has the major advantage to directly deal with the trade-off between the empirical loss and the KL divergence (\ie, the value of $C$ in Equation~\eqref{eq:thmpacbayes} is the one minimizing the bound).
\begin{theorem}
\label{thm:pacbayes}
\ul{Given a data independent prior distribution $\mathcal{B}^p$} and $\delta\in(0,1)$, with probability at least $1-\delta$ over a realization of the learning sample $S\sim\Dcal^n$, \underline{then, for all posterior $\Bcal$}: 
\begin{multline}
  \label{eq:thmpacbayes}
    \mathcal{L}_{\mathcal{D}}(F_{\mathcal{A}}) \leq \inf_{C>0} \bigg\{\tfrac{1}{1-e^{-C}} \left( 1 - \exp \left[ -C \hat{\mathcal{L}}_S(F_{\mathcal{A}}) -{} \right.\left.\tfrac{1}{n} \left(\KL(\mathcal{B} \| \mathcal{B}^p) + \ln \tfrac{2 \sqrt{n}}{\delta}\right) \right] \right) \bigg\}\,.
\end{multline}
\end{theorem}
A salient feature of the PAC-Bayesian bounds is that they are uniformly valid (with probability at least $1-\delta$) for the whole family of posterior. This is particularly suited for the design of a bound minimization algorithm, as the right-hand side of Equation~\eqref{eq:thmpacbayes} suggests an objective to minimize, and is providing a generalization guarantee even when the optimization procedure does not converge to a global minimum. 
Thus, we propose to train the ABNet architecture by minimizing the bound given in Theorem~\ref{thm:pacbayes} by stochastic gradient descent. That is, the following objective is optimized according to parameters $\Bcal$ and $C>0$:

\begin{equation}
\textstyle
\frac{1}{1-e^{-C}} \left( 1 - \exp \left[ -C \hat{\mathcal{L}}_S(F_{\mathcal{A}}) - \frac{1}{n}\left(\KL(\mathcal{B} \| \mathcal{B}^p) + \ln \tfrac{2 \sqrt{n}}{\delta} \right)\right] \right).
\label{eq:objfunc}
\end{equation}
Although this objective may appear similar to the ones of previous PAC-Bayes analyses, the proposed ABNet objective differs in two noticeable ways :\\
(1) The predictor output, and therefore the empirical loss of Equation~\eqref{eq:emploss}, corresponds to the exact BAM expectation and is computed thanks to the forward propagation routine of ABNet, instead of the more classical computational graph of Letarte et al. \cite{letarte2019dichotomize} or the approximation scheme of \citet{biggs2021}.\\
(2) The KL divergence of Equation~\eqref{eq:KL}, acting as a regularization term, does give the same penalty to weights of every layer in ABNet for networks of depth $L>2$. In contrast, the corresponding term in PBGNet \cite{letarte2019dichotomize} penalizes weights by a growing factor according to the layer depth.  



\section{Compacting the ABNet}
\label{section:compact}


Recall from Section \ref{section:dynamic_programming} that the aggregation output can be computed by the matrix product 
$\label{eq:ABNetMatricial}
    F_{\Acal(\Bcal)}(\xbf) \!=\! \begin{bmatrix}1,-1\end{bmatrix} \cdot (\Psibf_L (\Psibf_{L-1} ( \dots \mathbf{\Psi}_3( \mathbf{\Psi}_2 \Pbf_1(\xbf)) \dots))),$
with $\langle \Psibf_k \rangle_{k=2}^{L}$ and $\Pbf_1(\xbf)$ computed from parameters $\Bcal = \langle \Wbf_k \rangle_{k=1}^{L}$ according to Equations~\eqref{eq:psi_k} and~\eqref{eq:p1_x}.
From this point of view, ABNet simply computes a linear function of the \emph{leading hidden layer representation} $\Pbf_1(\xbf)$, highlighting a limitation of all binary (and discrete-valued) activated neural networks. Indeed, all matrices $\Psibf_k$ are solely based on the weights and do not rely on the input layer. Since there is no activation function between hidden layers, dot product associativity allows us to state the following. 
\begin{proposition}
\label{thm:2layers}
The output of an aggregation of BAMs $\Acal(\Bcal)$, where $\Bcal$ has leading hidden layer width of $d_1$ and arbitrary width for other hidden layers, can be obtained by forward propagating in a compact (with regard to depth) neural network having a leading hidden layer of width $2^{d_1}$ with $\Erf$ activation and an output layer of width $1$ with identity activation:
\begin{align} \label{eq:thm1}
F_{\mathcal{A}}(\mathbf{x}) = \Hbf \cdot \Pbf_1(\xbf) \,,
\end{align}
where $\Pbf_1(\xbf)$ is a vector of $2^{d_1}$ elements defining a probability distribution on the outputs of the leading hidden layer of $\Bcal$ on $\xbf$, and $\Hbf \in [-1,1]^{2^{d_1}}$ is a vector giving the expected output of the rest of the network given the output of the first layer, such that
\begin{align} \label{eq:multi_dot}
    \Hbf = \begin{bmatrix}1,-1\end{bmatrix} \cdot \mathbf{\Psi}_L \cdot \mathbf{\Psi}_{L-1} \cdot ... \cdot \mathbf{\Psi}_3 \cdot \mathbf{\Psi}_2\,.
\end{align} 
\end{proposition}
Since only $\Pbf_1(\xbf)$ changes in function of $\xbf$, for fixed weights one can numerically precompute $\Hbf \eqdots \big[h_{\sbf} \big]_{\sbf \in R_1}$ once and for all $\xbf$. In the underlying BAM this is analogous to precomputing the output for every representation outputted by the leading hidden layer. Every entry of~$\Hbf$ is a real number between $-1$ and $1$ since it represents an expectation on a BAM output. This observation leads us to the following corollary.

\begin{figure*}[t]
  \centering
 \includegraphics[trim={2mm 5mm 4mm 4mm},clip,width=0.8\linewidth]{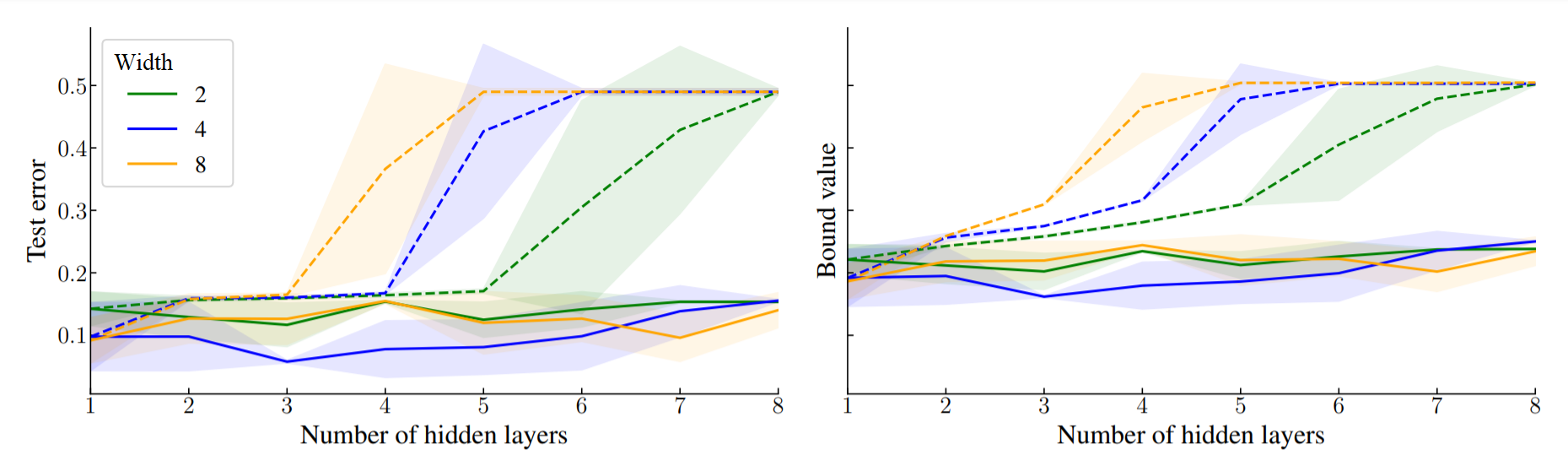}
  \caption{Impact of depth for \textbf{PBGNet (dashed)} and \textbf{ABNet (solid)} on test errors and bound values according to the width for mnistLH datasets.
  Results correspond to means and standard deviations over 5 repetitions.}
\label{fig:depth} 
\end{figure*}

\begin{corollary} \label{cor:shallow_output}
Notwithstanding the fact that the underlying BAM architecture can be arbitrarily deep, the aggregation output can always be expressed in the following shallow form, with $h_{\sbf} \in [-1,1]$:
\begin{align} \label{eq:shallow_form}
F_{\mathcal{A}}(\mathbf{x}) = \sum_{\sbf \in R_1} h_{\sbf} \prod_{i=1}^{d_1}  \bigg(\frac{1}{2} + \frac{\sbf^i}{2} \Erf\!\left(\frac{\mathbf{W}_{1}^i \cdot \mathbf{x}}{\sqrt{2} \lVert \mathbf{x} \rVert}\right)\bigg) \,.
\end{align}
Thus, forward propagation of ABNet can be computed in time constant with regard to $L$.
\end{corollary}
We call the algorithm that computes Equation~\eqref{eq:shallow_form} the \textit{Compact ABNet}. See Fig.~\ref{fig:compact_abnet_} for a visual representation. 
Interestingly, the PAC-Bayes generalization bound of Theorem~\ref{thm:pacbayes} is not obtainable directly from the Compact ABNet parameters. Therefore, our bound minimization algorithm requires the ABNet architecture. Noteworthy, our empirical experiments (Section~\ref{section:experiments}, Fig.~\ref{fig:depth}) show that training deeper ABNet can achieve better generalization than a shallower architecture, even when both share a Compact architecture of the same size. Compacting our stochastic version is also possible. Since the dot product must be executed on fixed $R'_k$s, the drawn samples must be predetermined and remain the same at each inference; this leads to a very concise classifier which performs just as well as the last learned Stochastic ABNet. As can be seen on Fig.~\ref{fig:varying_depth}, compact networks are much faster at inference time than their deep equivalent, as their complexity does not increase with depth. 

\begin{figure*}[t]
    \centering
    \begin{subfigure}[t]{0.98\textwidth}
        \centering
        \includegraphics[scale=0.9]{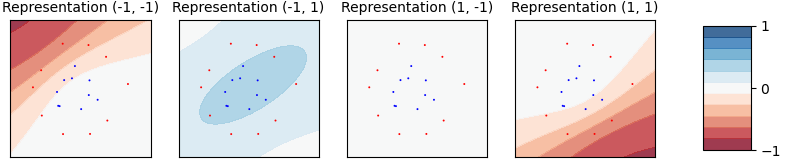}
        \caption{For all $\xbf$, the probabilities of outputting a specific representation at the leading hidden layer, obtained from $\Pbf_1(\xbf)$, multiplied by the expected output for each of these regions, given by $\Hbf$.}
        \label{fig:abnet_regions}
    \end{subfigure}
    \smallskip
    
     \begin{subfigure}[t]{0.3\textwidth}
        \centering
        \includegraphics[scale=0.4]{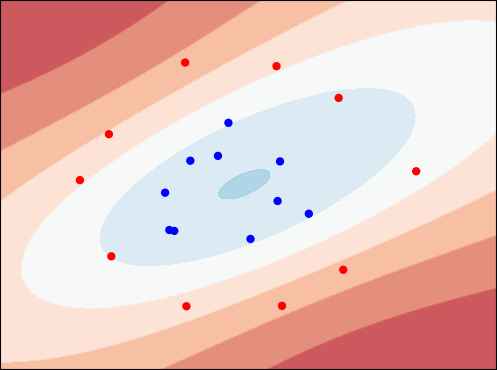}
        \caption{The aggregation output $F_{\Acal(\Bcal_M)}(\xbf)$ for all $\xbf$, obtained by summing over representations.}
        \label{fig:abnet_forward}
    \end{subfigure}
    \quad
     \begin{subfigure}[t]{0.33\textwidth}
        \centering
        \includegraphics[scale=0.4]{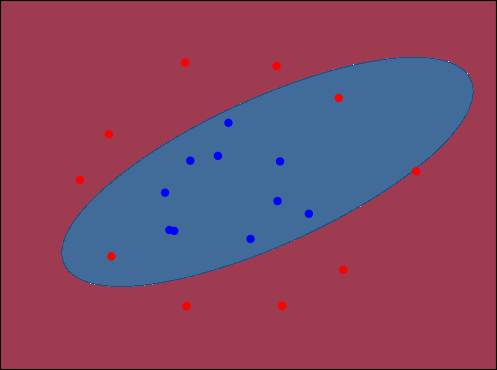}
        \caption{Using ABNet as a classifier, one can simply use $\sgn(F_{\Acal(\Bcal_M)}(\xbf))$ as the prediction for an input $\xbf$.}
        \label{fig:abnet_predict}
    \end{subfigure}
    \quad
     \begin{subfigure}[t]{0.3\textwidth}
        \centering
        \includegraphics[scale=0.4]{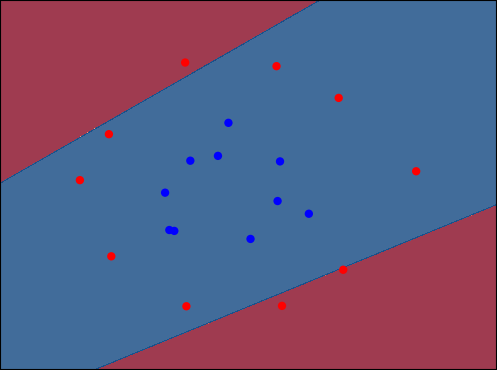}
        \caption{Using the BAM classifier from the MAP $F_{\Bcal_M}(\xbf)$, the decision boundary is less expressive.}
        \label{fig:bam_predict}
    \end{subfigure}
    \caption{Predictions of an ABNet and its underlying BAM with architecture $\langle 2,2,2,1 \rangle$, \ie, with two-dimensional inputs and two hidden layers of two neurons, on a toy dataset.}
    \label{fig:circles_dataset}
\end{figure*}

Our original approach of propagating probabilities over representations is what brings the light on the compactability phenomenon.
It is a well-known result that any function can be approximated to an arbitrary level of accuracy with a neural network having as few as \textit{one} hidden layer, given that the layer is wide enough \citep{hornik1989multilayer}. It has also been shown that a shallow "student" network can learn to mimic a deep "teacher" network to reach the same performance level \citep{caruana2014}. 
However, typical neural networks do not allow such an explicit construction that maps an initially (non-linear) deep structure to a shallow form. The result of Proposition \ref{thm:2layers} is a curiosity that is worth analyzing further.
Remarkably, there is a clear dichotomy between the roles of $\Pbf_1(\xbf)$ and $\Hbf$: the former transforms data points into a probability distribution over the leading hidden layer representations, whereas the latter gives the aggregation output for each of those representations. Put otherwise, the first layer serves as an embedding and the rest of the layers operate as a classifier. Fig.~\ref{fig:circles_dataset} illustrates the particularity of the prediction mechanism of ABNet. 

\textbf{The leading hidden layer.} \ 
Equation~\eqref{eq:thm1} implies that the leading hidden layer defines regions in the input space.
All subsequent hidden layers together express the output value of these regions. Fig.~\ref{fig:abnet_regions} shows how those regions divide the input space on the toy problem. 
Each region is associated with one of the four leading hidden layer binary representation.

Each neuron of the leading hidden layer of a BAM defines a hyperplane in the input space, where inputs on one side are mapped to $-1$, and $1$ on the other side.
Considering all regions that are enclosed between the hyperplanes yields up to $2^{d_1}$ regions, corresponding to the $2^{d_1}$ output representations $R_1$. Many of those regions may stray very far from the actual data. For example, in Fig.~\ref{fig:abnet_regions} the region corresponding to representation $(1, -1)$ exists on the other side of where the two planes meet, which is far from any existing data\footnote{By considering the few most important region, one could potentially \emph{interpret} ABNet predictions more easily than for classical neural networks. We consider this question as future work, and refer the reader to \citet{MontufarPCB14} for a study of regions in the broader context of neural networks with continuous activations.}.



\textbf{Additional hidden layers.} \ 
The vector $\Hbf$ represents by extension a function from $\{-1, 1\}^{d_1}$ to $[-1,1]$. Its role is to determine what sign should be outputted for each region defined by the leading hidden layer, with a confidence term. Its content is not arbitrary since it must be obtained from the weights of the subsequent hidden layers as in Equation~\eqref{eq:multi_dot}. Depth therefore adds expressivity to BAM aggregations by allowing regions created by the leading hidden layer to output uncorrelated signs, with more or less confidence.

As illustrated by Figures~\ref{fig:abnet_predict}-d, taking the output of ABNet is not equivalent as taking the output of its associated MAP. For the same parameters, the aggregation allows more complex regions than BAMs, taking advantage that an input can belong to several regions, with certain probabilities; there exist points $\xbf$ and parameters $\Bcal_M$ for which
$\sgn (F_{\Acal(\Bcal_M)}(\xbf))\, {\neq}\, F_{\Bcal_M}(\xbf)$. For instance, many incorrectly classified $-1$ data points in Fig.~\ref{fig:bam_predict} fall within the correct region in Fig.~\ref{fig:abnet_predict} because ABNet can compensate the proximity to the central $+1$ region with a lesser proximity with \textit{two} $-1$ regions. 
It is therefore worthy to use the aggregation as a predictor by itself instead of its MAP, for its expressive power. Indeed, \mbox{\citet{ZhuDS19}} observe that ensemble methods on binary neural networks confer stability with regard to input and parameter perturbations, which leads to better generalization.

\section{Numerical experiments}
\label{section:experiments}
We evaluated our proposed approach ABNet by following the experimental framework of \citet{letarte2019dichotomize}, on the same six binary classification datasets: \textit{ads} and \textit{adult} from the UCI repository \citep{Dua:2019}, along with four MNIST \citep{lecun1998gradient} binary variants \textit{mnistLH}  (labels $\{0,1,2,3,4\}$ form the "Low" class, and $\{5,6,7,8,9\}$ the "High" class), \textit{mnist17}, \textit{mnist49} and \textit{mnist56} (only examples labeled respectively $1\&7$, $4\&9$, and $5\&6$ are retained).\footnote{We conducted PBGNet experiments using the \href{https://github.com/gletarte/dichotomize-and-generalize}{publicly available source code} of Letarte et al. \cite{letarte2019dichotomize}. We did not perform experiments using the approximation scheme of \citet{biggs2021} as their code has not been made public. Moreover, the experiments of the latter were performed solely on \textit{mnistLH} dataset, showing similar accuracy than PBGNet for a similar architecture.}
As the exact versions of PBGNet and ABNet are limited by their exponential complexity regarding their width, we explored narrow network architectures (widths $d \in \{2, 4, 8\}$) and wider architectures (widths $d \in \{10, 50, 100\}$) accessible only to stochastic versions, all for $1$ to $3$ hidden layers. All experiments were repeated with 5 different random train/test dataset splits and weights initializations. Networks parameters are optimized using Adam \citep{kingma2014adam} for the following learning rate values: $\{0.1, 0.01, 0.001, 0.0001\}$. Training is performed for 100 epochs with early stopping after 20 epochs without improvement.

\renewcommand{\arraystretch}{0.8} 
\begin{table}[t]
    \centering
    \caption{Experiment results. On narrow architectures (left column), standard versions of PBGNet and ABNet are used, while their stochastic versions are used on wide architectures (right column). For each dataset and model, the best-performing set of parameters over the repetitions is retained. Shown are the number of hidden layers ($L{-}1$), hidden size ($d$), bound value and error rate on the train data (Error$_{S}$)  and on the test data for the model (Error$_{T}$) and the associated MAP BAM. The standard deviation over the five repetitions is shown for the test error. For both BC and BNN algorithms, the Error$_T$ and MAP columns share the same results, as these models don't rely on aggregation.
    }
    \label{tab:selected_models_overview}
    \setlength{\tabcolsep}{2.2pt}
    {\begin{tabular}{ll|cccccc|cccccc}
    \toprule
    & \scriptsize  & \multicolumn{6}{c|}{\scriptsize Narrow architectures} & \multicolumn{6}{c}{\scriptsize Wide architectures} \\
    \scriptsize Dataset & \scriptsize  Model & \scriptsize  $L{-}1$ & \scriptsize   $d$ & \scriptsize  Bound & \scriptsize  Error$_{S}$ & \scriptsize  Error$_{T}$ & \scriptsize  MAP & \scriptsize  $L{-}1$ & \scriptsize   $d$ & \scriptsize  Bound & \scriptsize  Error$_{S}$ & \scriptsize  Error$_{T}$ & \scriptsize  MAP\\
    \midrule
    \scriptsize \multirow{7}{*}{ads} 
    & \scriptsize  \scriptsize PBGNet & \scriptsize   3 & \scriptsize   2 & \scriptsize   0.192 & \scriptsize   0.140 & \scriptsize   0.141 $\pm$~0.012 & \scriptsize   0.141 & \scriptsize   3 & \scriptsize  10 & \scriptsize  0.213 & \scriptsize  0.140 & \scriptsize  0.141 $\pm$~0.010 & \scriptsize  0.141 \\
    & \scriptsize  ABNet & \scriptsize   3 & \scriptsize   2 & \scriptsize   0.192 & \scriptsize   0.140 & \scriptsize   0.141 $\pm$~0.012 & \scriptsize   0.141 & \scriptsize   2 & \scriptsize  10 & \scriptsize  0.216 & \scriptsize  0.140 & \scriptsize  0.141 $\pm$~0.010 & \scriptsize  0.141\\
    & \scriptsize  PBGNet$_{\ell}$ & \scriptsize   3 & \scriptsize   4 & \scriptsize   1.000 & \scriptsize   0.018 & \scriptsize   0.026 $\pm$~0.004 & \scriptsize   0.027 & \scriptsize   3 & \scriptsize  10 & \scriptsize  1.000 & \scriptsize  0.020 & \scriptsize  0.026 $\pm$~0.006 & \scriptsize  0.028 \\
    & \scriptsize  ABNet$_{\ell}$ & \scriptsize   3 & \scriptsize   4 & \scriptsize   0.887 & \scriptsize   0.015 & \scriptsize   0.026 $\pm$~0.003 & \scriptsize   0.026 & \scriptsize   2 & \scriptsize  50 & \scriptsize  1.000 & \scriptsize  0.020 & \scriptsize  0.026 $\pm$~0.005 & \scriptsize  0.025 \\
    & \scriptsize  EBP & \scriptsize   2 & \scriptsize   2 & \scriptsize  -- & \scriptsize  0.003 & \scriptsize   0.040 $\pm$~0.008 & \scriptsize  0.054 & \scriptsize   3 & \scriptsize  10 & \scriptsize  -- & \scriptsize  0.005 & \scriptsize  0.035 $\pm$~0.006 & \scriptsize  0.049  \\
    & \scriptsize  BC & \scriptsize  1 & \scriptsize  4 & \scriptsize  -- & \scriptsize  0.025 & \scriptsize   0.031 $\pm$ 0.004 & \scriptsize   \textcolor{lightgray}{0.031} & \scriptsize  1 & \scriptsize  10 & \scriptsize  -- & \scriptsize  0.021 & \scriptsize  0.032 $\pm$ 0.005 & \scriptsize  \textcolor{lightgray}{0.032} \\
    & \scriptsize  BNN & \scriptsize   1 & \scriptsize   8 & \scriptsize  -- & \scriptsize  0.037 & \scriptsize  0.038 $\pm$ 0.004 & \scriptsize  \textcolor{lightgray}{0.038} & \scriptsize   1 & \scriptsize  100 & \scriptsize  -- & \scriptsize  0.029 & \scriptsize  0.032 $\pm$ 0.005 & \scriptsize  \textcolor{lightgray}{0.032} \\
    \midrule
    \scriptsize \multirow{7}{*}{adult} 
    & \scriptsize  PBGNet & \scriptsize   1 & \scriptsize   2 & \scriptsize   0.208 & \scriptsize   0.157 & \scriptsize   0.160 $\pm$~0.003 & \scriptsize   0.158 & \scriptsize   1 & \scriptsize  10 & \scriptsize  0.216 & \scriptsize  0.156 & \scriptsize  0.159 $\pm$~0.002 & \scriptsize  0.158 \\
    & \scriptsize  ABNet & \scriptsize   1 & \scriptsize   2 & \scriptsize   0.208 & \scriptsize   0.157 & \scriptsize   0.160 $\pm$~0.003 & \scriptsize   0.158 & \scriptsize   1 & \scriptsize  10 & \scriptsize  0.216 & \scriptsize  0.156 & \scriptsize  0.160 $\pm$~0.002 & \scriptsize  0.158 \\ 
    & \scriptsize  PBGNet$_{\ell}$ & \scriptsize   3 & \scriptsize   4 & \scriptsize   0.723 & \scriptsize   0.135 & \scriptsize   0.149 $\pm$~0.002 & \scriptsize   0.156 & \scriptsize   2 & \scriptsize  10 & \scriptsize  0.360 & \scriptsize  0.146 & \scriptsize  0.151 $\pm$~0.002 & \scriptsize  0.164\\
    & \scriptsize  ABNet$_{\ell}$ & \scriptsize   3 & \scriptsize   4 & \scriptsize   0.780 & \scriptsize   0.132 & \scriptsize   0.149 $\pm$~0.003 & \scriptsize   0.150 & \scriptsize   3 & \scriptsize  10 & \scriptsize  0.541 & \scriptsize  0.143 & \scriptsize  0.151 $\pm$~0.002 & \scriptsize  0.151 \\
    & \scriptsize  EBP   & \scriptsize  1 & \scriptsize  8 & \scriptsize  -- & \scriptsize  0.145 & \scriptsize  0.152 $\pm$~0.003 & \scriptsize  0.166 & \scriptsize  2 & \scriptsize  100 & \scriptsize  -- & \scriptsize  0.049 & \scriptsize  0.186 $\pm$~0.001 & \scriptsize  0.189  \\
    & \scriptsize  BC    & \scriptsize  1 & \scriptsize  8 & \scriptsize  -- & \scriptsize  0.142 & \scriptsize  0.151 $\pm$ 0.002 & \scriptsize  \textcolor{lightgray}{0.151} & \scriptsize  1 & \scriptsize  50 & \scriptsize  -- & \scriptsize  0.160 & \scriptsize  0.164 $\pm$ 0.001 & \scriptsize  \textcolor{lightgray}{0.164} \\
    & \scriptsize  BNN   & \scriptsize  1 & \scriptsize  2 & \scriptsize  -- & \scriptsize  0.180 & \scriptsize  0.182 $\pm$ 0.017 & \scriptsize  \textcolor{lightgray}{0.182} & \scriptsize  1 & \scriptsize  100 & \scriptsize  -- & \scriptsize  0.157& \scriptsize  0.165 $\pm$ 0.002 & \scriptsize  \textcolor{lightgray}{0.165} \\
    \midrule
    \scriptsize\multirow{7}{*}{mnist17} 
    & \scriptsize  PBGNet & \scriptsize   1 & \scriptsize   2 & \scriptsize   0.036 & \scriptsize   0.005 & \scriptsize   0.006 $\pm$~0.001 & \scriptsize   0.006 & \scriptsize   1 & \scriptsize  10 & \scriptsize  0.041 & \scriptsize  0.005 & \scriptsize  0.006 $\pm$~0.001 & \scriptsize  0.006 \\
    & \scriptsize  ABNet & \scriptsize   1 & \scriptsize   2 & \scriptsize   0.036 & \scriptsize   0.005 & \scriptsize   0.006 $\pm$~0.001 & \scriptsize   0.006 & \scriptsize   1 & \scriptsize  10 & \scriptsize  0.041 & \scriptsize  0.005 & \scriptsize  0.006 $\pm$~0.001 & \scriptsize  0.006 \\
    & \scriptsize  PBGNet$_{\ell}$ & \scriptsize   3 & \scriptsize   4 & \scriptsize   1.000 & \scriptsize   0.002 & \scriptsize   0.005 $\pm$~0.001 & \scriptsize   0.005 & \scriptsize   2 & \scriptsize  10 & \scriptsize  1.000 & \scriptsize  0.002 & \scriptsize  0.005 $\pm$~0.001 & \scriptsize  0.006 \\
    & \scriptsize  ABNet$_{\ell}$ & \scriptsize   3 & \scriptsize   2 & \scriptsize   0.829 & \scriptsize   0.001 & \scriptsize   0.004 $\pm$~0.001 & \scriptsize   0.004 & \scriptsize   3 & \scriptsize  10 & \scriptsize  0.607 & \scriptsize  0.002 & \scriptsize  0.005 $\pm$~0.001 & \scriptsize  0.005 \\
    & \scriptsize  EBP & \scriptsize  1 & \scriptsize   2 & \scriptsize  -- & \scriptsize   0.000 & \scriptsize  0.006 $\pm$~0.001 & \scriptsize   0.006 & \scriptsize  2 & \scriptsize  10 & \scriptsize  -- & \scriptsize  0.000 & \scriptsize  0.005 $\pm$~0.000 & \scriptsize   0.005 \\
    & \scriptsize  BC & \scriptsize  2 & \scriptsize  4 & \scriptsize  -- & \scriptsize   0.004 & \scriptsize  0.010 $\pm$ 0.002 & \scriptsize  \textcolor{lightgray}{0.010} & \scriptsize  3 & \scriptsize  50 & \scriptsize  -- & \scriptsize  0.003 & \scriptsize  0.006 $\pm$ 0.001 & \scriptsize  \textcolor{lightgray}{0.006} \\
    & \scriptsize  BNN & \scriptsize  1 & \scriptsize  8 & \scriptsize  -- & \scriptsize  0.003 & \scriptsize  0.008 $\pm$ 0.001 & \scriptsize  \textcolor{lightgray}{0.008} & \scriptsize  1 & \scriptsize  100 & \scriptsize  -- & \scriptsize  0.004 & \scriptsize  0.007 $\pm$ 0.001 & \scriptsize  \textcolor{lightgray}{0.007} \\ 
    \midrule
    \scriptsize\multirow{7}{*}{mnist49} 
    & \scriptsize  PBGNet & \scriptsize   1 & \scriptsize   2 & \scriptsize   0.136 & \scriptsize   0.036 & \scriptsize   0.036 $\pm$~0.004 & \scriptsize   0.036 & \scriptsize  1 & \scriptsize  10 & \scriptsize  0.149 & \scriptsize  0.037 & \scriptsize  0.037 $\pm$~0.004 & \scriptsize  0.036 \\
    & \scriptsize  ABNet & \scriptsize   1 & \scriptsize   2 & \scriptsize   0.136 & \scriptsize   0.036 & \scriptsize   0.036 $\pm$~0.004 & \scriptsize   0.036 & \scriptsize   1 & \scriptsize  10 & \scriptsize  0.147 & \scriptsize  0.038 & \scriptsize  0.037 $\pm$~0.004 & \scriptsize  0.036 \\
    & \scriptsize  PBGNet$_{\ell}$ & \scriptsize   2 & \scriptsize   4 & \scriptsize   1.000 & \scriptsize   0.008 & \scriptsize   0.020 $\pm$~0.004 & \scriptsize   0.020 & \scriptsize   1 & \scriptsize  50 & \scriptsize  0.992 & \scriptsize  0.004 & \scriptsize  0.012 $\pm$~0.003 & \scriptsize  0.012 \\
    & \scriptsize  ABNet$_{\ell}$ & \scriptsize   3 & \scriptsize   8 & \scriptsize   1.000 & \scriptsize   0.004 & \scriptsize   0.017 $\pm$~0.003 & \scriptsize  0.017 & \scriptsize  3 & \scriptsize  10 & \scriptsize  1.000 & \scriptsize  0.024 & \scriptsize  0.029 $\pm$~0.003 & \scriptsize  0.027 \\
    & \scriptsize  EBP   & \scriptsize  3 & \scriptsize  8 & \scriptsize  -- & \scriptsize  0.020 & \scriptsize  0.033 $\pm$~0.003 & \scriptsize  0.034 & \scriptsize  2 & \scriptsize  10 & \scriptsize  -- & \scriptsize  0.001 & \scriptsize  0.021 $\pm$~0.004 & \scriptsize  0.026\\
    & \scriptsize  BC    & \scriptsize  2 & \scriptsize  8 & \scriptsize  -- & \scriptsize  0.007 & \scriptsize  0.016 $\pm$ 0.002 & \scriptsize  \textcolor{lightgray}{0.016} & \scriptsize  1 & \scriptsize  100 & \scriptsize  -- & \scriptsize  0.005 & \scriptsize  0.015 $\pm$ 0.003 & \scriptsize  \textcolor{lightgray}{0.015}\\
    & \scriptsize  BNN   & \scriptsize  1 & \scriptsize  2 & \scriptsize  -- & \scriptsize  0.030 & \scriptsize  0.037 $\pm$ 0.003 & \scriptsize  \textcolor{lightgray}{0.037} & \scriptsize  1 & \scriptsize  100 & \scriptsize  -- & \scriptsize  0.011 & \scriptsize  0.023 $\pm$ 0.003 & \scriptsize  \textcolor{lightgray}{0.023} \\
    \midrule
    \scriptsize\multirow{7}{*}{mnist56} 
    & \scriptsize  PBGNet & \scriptsize   1 & \scriptsize   2 & \scriptsize   0.084 & \scriptsize   0.021 & \scriptsize   0.023 $\pm$~0.006 & \scriptsize   0.023 & \scriptsize   1 & \scriptsize  10 & \scriptsize  0.090 & \scriptsize  0.023 & \scriptsize  0.025 $\pm$~0.003 & \scriptsize  0.024 \\
    & \scriptsize  ABNet & \scriptsize   1 & \scriptsize   2 & \scriptsize   0.084 & \scriptsize   0.021 & \scriptsize   0.023 $\pm$~0.006 & \scriptsize   0.023 & \scriptsize   1 & \scriptsize  10 & \scriptsize  0.090 & \scriptsize  0.023 & \scriptsize  0.025 $\pm$~0.003 & \scriptsize  0.024 \\
    & \scriptsize  PBGNet$_{\ell}$ & \scriptsize   2 & \scriptsize   8 & \scriptsize   1.000 & \scriptsize   0.004 & \scriptsize   0.011 $\pm$~0.003 & \scriptsize   0.011 & \scriptsize   1 & \scriptsize  50 & \scriptsize  0.974 & \scriptsize  0.003 & \scriptsize  0.008 $\pm$~0.002 & \scriptsize  0.008 \\
    & \scriptsize  ABNet$_{\ell}$ & \scriptsize  3 & \scriptsize   8 & \scriptsize   0.999 & \scriptsize   0.004 & \scriptsize   0.009 $\pm$~0.002 & \scriptsize   0.009 & \scriptsize  1 & \scriptsize  10 & \scriptsize  1.000 & \scriptsize  0.010 & \scriptsize  0.017 $\pm$~0.002 & \scriptsize  0.016 \\
    & \scriptsize  EBP & \scriptsize  3 & \scriptsize  8 & \scriptsize  -- & \scriptsize  0.001 & \scriptsize  0.010 $\pm$~0.004 & \scriptsize  0.015 & \scriptsize  2 & \scriptsize  10 & \scriptsize  -- & \scriptsize  0.000 & \scriptsize  0.019 $\pm$~0.004 & \scriptsize  0.021 \\
    & \scriptsize  BC & \scriptsize  1 & \scriptsize  8 & \scriptsize  -- & \scriptsize  0.002 & \scriptsize  0.009 $\pm$ 0.004 & \scriptsize  \textcolor{lightgray}{0.009} & \scriptsize  3 & \scriptsize  50 & \scriptsize  -- & \scriptsize  0.004 & \scriptsize  0.010 $\pm$ 0.003 & \scriptsize  \textcolor{lightgray}{0.010} \\
    & \scriptsize  BNN & \scriptsize  1 & \scriptsize  8 & \scriptsize  -- & \scriptsize  0.013 & \scriptsize  0.023 $\pm$ 0.003 & \scriptsize  \textcolor{lightgray}{0.023} & \scriptsize  1 & \scriptsize  100 & \scriptsize  -- & \scriptsize  0.004 & \scriptsize  0.012 $\pm$ 0.001 & \scriptsize  \textcolor{lightgray}{0.012} \\
    \midrule
    \scriptsize\multirow{7}{*}{mnistLH} 
    & \scriptsize  PBGNet & \scriptsize   1 & \scriptsize   8 & \scriptsize   0.186 & \scriptsize   0.091 & \scriptsize   0.092 $\pm$~0.036 & \scriptsize   0.093 & \scriptsize   1 & \scriptsize  10 & \scriptsize  0.167 & \scriptsize  0.058 & \scriptsize  0.059 $\pm$~0.010 & \scriptsize  0.060 \\
    & \scriptsize  ABNet & \scriptsize   3 & \scriptsize   4 & \scriptsize   0.162 & \scriptsize   0.056 & \scriptsize   0.058 $\pm$~0.002 & \scriptsize   0.059 & \scriptsize   2 & \scriptsize  10 & \scriptsize  0.187 & \scriptsize  0.087 & \scriptsize  0.088 $\pm$~0.006 & \scriptsize  0.087 \\
    & \scriptsize   PBGNet$_{\ell}$ & \scriptsize   3 & \scriptsize   8 & \scriptsize   1.000 & \scriptsize   0.018 & \scriptsize   0.038 $\pm$~0.002 & \scriptsize   0.047 & \scriptsize   1 & \scriptsize  100 & \scriptsize  0.998 & \scriptsize  0.006 & \scriptsize  0.022 $\pm$~0.001 & \scriptsize  0.024 \\
    & \scriptsize  ABNet$_{\ell}$ & \scriptsize   2 & \scriptsize   8 & \scriptsize   0.998 & \scriptsize   0.025 & \scriptsize   0.042 $\pm$~0.006 & \scriptsize   0.043 & \scriptsize   3 & \scriptsize  10 & \scriptsize  0.895 & \scriptsize  0.050 & \scriptsize  0.060 $\pm$~0.005 & \scriptsize  0.058 \\
    & \scriptsize  EBP   & \scriptsize  3 & \scriptsize  8 & \scriptsize  -- & \scriptsize  0.016 & \scriptsize  0.043 $\pm$~0.002 & \scriptsize  0.082 & \scriptsize  1 & \scriptsize  100 & \scriptsize  -- & \scriptsize  0.001 & \scriptsize  0.027 $\pm$~0.001 & \scriptsize  0.032  \\
    & \scriptsize  BC    & \scriptsize  2 & \scriptsize  8 & \scriptsize  -- & \scriptsize  0.023 & \scriptsize  0.035 $\pm$ 0.001 & \scriptsize  \textcolor{lightgray}{0.035} & \scriptsize  1 & \scriptsize  100 & \scriptsize  -- & \scriptsize  0.013 & \scriptsize  0.027 $\pm$ 0.001 & \scriptsize  \textcolor{lightgray}{0.027} \\
    & \scriptsize  BNN   & \scriptsize  1 & \scriptsize  2 & \scriptsize  -- & \scriptsize  0.123 & \scriptsize  0.133 $\pm$ 0.004 & \scriptsize  \textcolor{lightgray}{0.133} & \scriptsize  1 & \scriptsize  100 & \scriptsize  -- & \scriptsize  0.023 & \scriptsize  0.036 $\pm$ 0.001 & \scriptsize  \textcolor{lightgray}{0.036} \\
    \bottomrule
    \end{tabular}
    }
\end{table}

We first compare ABNet to its direct counterpart PBGNet  \citep{letarte2019dichotomize}, both directly optimizing the PAC-Bayesian generalization bound during learning, with the prior distribution defined by the network weights random initialization.   
We also explore the minimization of the empirical loss with the variants ABNet$_{\ell}$ (Eq.~\ref{eq:emploss}) and PBGNet$_{\ell}$, where 20\% of the training data is used as a validation set for model selection.

Even if our primary focus is on the learning of a BAM aggregation, the optimization procedure of PBGNet and ABNet may be used to learn a single BAM, as it is not itself learnable with standard gradient descent methods. We thus compare the \emph{Maximum-A-Posteriori} (MAP) networks of both aggregated methods to three algorithms of the literature for learning neural networks with binary weights and/or activations: \textit{Expectation Backpropagation} \citep{soudry2014expectation} (EBP) with real-valued weights and binary activations, 
\textit{Binarized Neural Network} \citep{hubara2016} (BNN) 
with both binary weights and activations, 
and \textit{BinaryConnect} \citep{courbariaux2015binaryconnect} (BC) 
with binary weights but ReLU activations.
Experiments involving EBP, BNN or BC are performed using fully connected networks, following the procedure used for ABNet$_{\ell}$ and PBGNet$_{\ell}$.  

\textbf{Narrow architectures.} \ 
The PAC-Bayesian inspired models with empirical loss minimization (PBGNet$_{\ell}$ and ABNet$_{\ell}$) obtain competitive error rates  (similar to the results achieved by BC using ReLU activations and binary weights). However, the empirical loss minimization procedure lead to non-informative generalization bounds values. 
When considering the bound for optimization and model selection for PBGNet and ABNet, selected network architectures are smaller with usually a single hidden layer, as the objective function contains a regularization term on the weight values (see Eq.~\ref{eq:objfunc}), and the error rates grow while bound values improve to a relevant level. Also, bound minimization algorithms are far less prone to \emph{overfitting} than traditional optimization schemes, as their training errors are remarkably close to their testing errors (recall that PBGNet and ABNet are equivalent for one hidden layer).
On the larger and harder dataset mnistLH, the narrow ABNet achieves better error rate and bound value than PBGNet by selecting a deeper architecture thanks to its less penalizing KL divergence regularization.
On the performances of the MAP induced BAM networks, error rates are usually similar or slightly higher than their aggregated counterpart, implying these approaches are suitable algorithms to learn BAM networks. 
%

\textbf{Wide architectures.} \ 
For all algorithms and most datasets, obtained results for wide and narrow binary neural networks are surprisingly similar. This reveals that constraining ABNet's width to compute the exact aggregation output is not a major caveat. In particular, when one seeks  tight PAC-Bayesian guarantees, lower complexity of narrow models should be favored.
That being said, the proposed stochastic training for ABNet enables scaling to wider networks. While achieving most of the time comparable results to the stochastic PBGNet, the obtained risk on the large mnistLH dataset suggests that the approximation scheme of ABNet may not be as effective as the exact computation.

\textbf{Deep architectures.} \ 
A key improvement of ABNet over PBGNet is the KL divergence computation which is not hindered by a growing factor penalizing the weights according to the network depth.
This property should allow ABNet to learn deeper networks with tighter generalization bounds, which we investigated on the mnistLH dataset by extending the main experiment up to 8 hidden layers. 
Results are presented in Figure \ref{fig:depth} where the difference of behaviour between the models is clearly highlighted.
For a small number of hidden layers, the performances are similar, but as the number of hidden layer grows, bound values for PBGNet sharply rise and test error rates degrade significantly.
On the other hand, bound values are relatively stable for ABNet, indicating its potential to learn deep neural network architectures (the minimum bound is achieved for 3 hidden layers of width 4, which adequately indicates the best test error).



\section{Conclusion}
\label{section:conclusion}

Many desirable properties stem from considering a PAC-Bayesian analysis of aggregations over binary activated networks. 
Like the previous approaches on which we build \citep{letarte2019dichotomize,biggs2021}, our proposed learning algorithm gives a sensible way to optimize the parameters of such networks, and provides tight bounds on the generalization error of deep architectures.
The originality of our work lies in the focus on the exact computation of aggregation of narrow networks, for which we derive an atypical training scheme based on the propagation of probabilities over hidden layer representations.
We further extend the analysis to expose the dichotomy between the first layers versus the others. We believe the latter observation is a sensible tool to understand the expressivity conferred by a network's architecture. Pursuing this research direction could contribute to a line of work \cite{le2010deep,SzymanskiM12a,KidgerL20} studying the role of depth in neural networks, but in the context of model aggregation. An interesting perspective is to consider more complex distributions over the network parameters.


\printbibliography[heading=subbibintoc]




\end{document}